%% file: main.tex
\documentclass[twocolumn,10pt]{article}

\usepackage[margin=0.75in, columnsep=0.25in]{geometry}
\usepackage[T1]{fontenc}
\usepackage[utf8]{inputenc}
\usepackage{lmodern}
\usepackage{microtype}
\usepackage{amsmath,amssymb,amsthm}
\usepackage{booktabs}
\usepackage{array}
\usepackage{xcolor}
\usepackage{tikz}
\usepackage{pgfplots}
\pgfplotsset{compat=1.18}
\usepackage{cite}
\usepackage{hyperref}
\hypersetup{
  colorlinks=true,
  linkcolor=teal!70!black,
  citecolor=blue!70!black,
  urlcolor=blue!70!black
}

\title{\textbf{Grammar-as-a-Drafter: More Tokens per Speculative Decoding Step\\on Grammar-Constrained Structured Outputs}}

\author{
  John Zhang$^{*}$ \qquad Silky Arora\\
  \small Google\\
  \small $^{*}$Corresponding author: \texttt{johnqiangzhang@google.com}\\
  \small Implementation: \url{https://github.com/vllm-project/vllm-torchtpu/pull/3321}
}

\date{\today}

\begin{document}
\maketitle

\begin{abstract}
Speculative decoding speeds up generation by letting a lightweight drafter propose tokens that the target model verifies in one forward pass. Under structured outputs (JSON schemas, regular expressions, tool-call formats) the drafter works against a constraint it does not know: its proposals for keys, delimiters and indentation are often rejected or invalid. Meanwhile the grammar engine knows exactly which tokens must come next.

We present \textbf{Grammar-as-a-Drafter (GaaD)}, a technique that improves speculative decoding for structured outputs by making the grammar a second drafter. Before each verify step, the scheduler merges the drafter's proposal with grammar-forced tokens (tokens uniquely determined by the grammar, i.e., out-degree 1 in the grammar automaton) into one draft row of width $K$ (which can exceed the drafter's proposal length $K_M$): proposed tokens that match the forced token are promoted to \texttt{GRAMMAR}, the proposal is truncated where it disagrees, and grammar tokens fill the remaining slots. The verify step accepts \texttt{GRAMMAR} drafts without computing their logits, so GaaD composes with any drafter and adds no kernel and no compilation on the critical path. Accepting a \texttt{GRAMMAR} draft with probability 1 gives up the lossless guarantee of standard speculative decoding: every output still satisfies the grammar, but the grammar fixes text, not tokens, so committing the canonical tokenization of forced text changes the output distribution at and after \texttt{GRAMMAR} slots relative to grammar-masked decoding of the target.

We implement GaaD in vLLM on TPU (\texttt{vllm-torchtpu}) and in an independent serving runtime written in Rust, and evaluate it on TPU v7x with MTP, DFlash and n-gram drafters on a JSON character-profile benchmark. GaaD raises the average acceptance length (tokens per verify step, including the bonus token) from 2.06 to 5.43 for MTP on Qwen3.5-0.8B (1.99 to 5.31 on Qwen3.5-35B-A3B), from 1.57 to 4.13 for DFlash and from 1.30 to 3.64 for n-gram on Qwen3-4B, and decodes 2.66--2.81$\times$ faster than the best configuration of each drafter alone at batch size 1 and 1.14--1.70$\times$ faster at batch size 64. In the Rust runtime, MTP with GaaD is 2.23$\times$ faster than MTP alone on Qwen3.5-0.8B and 2.02$\times$ on Qwen3.5-397B-A17B-FP8, and still 1.22$\times$ faster at 128 concurrent requests. Under mixed traffic, with a quarter to half of the requests structured, GaaD finishes the workload 1.03--1.23$\times$ sooner and each structured request 2.2--3.2$\times$ sooner in \texttt{vllm-torchtpu}, while the plain-text requests that share a batch with them see a 7--42\% higher time per output token. We also find that grammar-masked greedy decoding can choose a non-canonical tokenization of forced text, so committing grammar drafts can yield a different, equally valid output; under an fp32 reference model, neither output is systematically more likely.
\end{abstract}

\section{Introduction}

Structured outputs are now routine in LLM serving: function calls, JSON extraction and agent actions are generated under a grammar that the engine enforces with a token mask~\cite{willard2023efficient,ugare2024syncode,dong2025xgrammar}. Speculative decoding is equally routine: a drafter---an n-gram table~\cite{saxena2023promptlookup}, an MTP head~\cite{gloeckle2024better}, EAGLE~\cite{li2024eagle,li2025eagle3}, or a block-diffusion model such as DFlash~\cite{dflash}---proposes several tokens, and the target model verifies them in one forward pass~\cite{leviathan2023fast,chen2023accelerating}. Engines combine the two by masking the target's logits at every draft position with the grammar's bitmask.

The combination leaves performance on the table. The drafter does not see the grammar, so on the deterministic parts of a structured output---keys, quotes, colons, indentation---it often proposes tokens the grammar forbids or spells the text differently, and those drafts are rejected. At the same time, the grammar engine knows the next tokens with certainty: in the JSON format we evaluate (Section~\ref{sec:eval}), which has ten value fields, every key and delimiter is fixed and only the values are chosen by the model. Each step thus pays for a draft that is partly wrong while the right tokens are available on the host.

\textbf{Grammar-as-a-Drafter (GaaD)} closes this gap by treating the grammar as a second drafter. For each request and step, the scheduler builds one draft row from the drafter's proposal and the grammar. Drafter tokens that match the forced tokens are kept, the proposal is cut where it leaves the forced path, and forced tokens fill the rest of the row (Figure~\ref{fig:planner}). Every slot carries its provenance, \texttt{MODEL} or \texttt{GRAMMAR}. The verify step that the engine already runs processes the row, with one change: it accepts \texttt{GRAMMAR} slots.

The verify step accepts \texttt{GRAMMAR} drafts without comparing them with the target, so it need not compute their logits at all, and the row can be wider than the drafter's proposal: the drafter fills its usual few slots and the grammar the rest. Accepting grammar drafts is attractive---the grammar admits no other text---but not entirely neutral: the grammar fixes text, not tokenization, and committing the canonical tokenization of a forced string can change the tokens, and with them the values, that follow (Section~\ref{sec:semantics}). GaaD therefore gives up the lossless guarantee of standard speculative decoding, which reproduces the target's output distribution exactly: \texttt{MODEL} slots are verified as usual, but a \texttt{GRAMMAR} slot is accepted with probability 1, so the distribution at and after it differs from grammar-masked decoding of the target, under sampling and, through tokenization drift, under greedy decoding as well. Under sampling, dropping a drafted token the grammar forbids before verification, which vLLM's structured-output path already does without GaaD, also shifts the distribution at free slots (Section~\ref{sec:limits}). Section~\ref{sec:equivalence} measures what this changes in the outputs, and Section~\ref{sec:limits} discusses the quality impact.

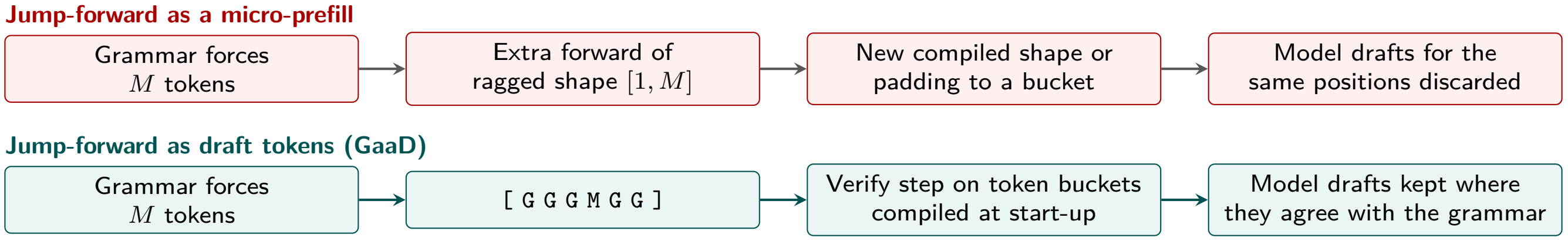
\begin{figure*}[t]
  \centering
  \resizebox{\textwidth}{!}{\input{figures/fig1_conflicts.tex}}
  \caption{Forced tokens as a separate prefill need a new shape per jump length and compete with drafts for the same positions. GaaD writes them into the draft row, which the verify step already handles.}
  \label{fig:conflicts}
\end{figure*}

GaaD also subsumes jump-forward decoding~\cite{yin2024compressedfsm,zheng2024sglang}, which appends forced strings with a separate variable-length prefill. On an engine that compiles shapes ahead of time, such as vLLM on TPU, that prefill requires either extra compiled shapes or padding an additional forward pass to a precompiled bucket, and it discards the drafter's proposals for the same positions (Figure~\ref{fig:conflicts}). Writing forced strings as draft tokens avoids both: it reuses the verify step's precompiled buckets with no extra forward pass and merges with the drafter's proposal instead of discarding it.

This paper makes four contributions:
\begin{enumerate}
  \item \textbf{GaaD}, a drafting technique that merges grammar-forced tokens with any drafter's proposal in one draft row, with per-slot provenance. It is lightweight: it needs no new kernel and only a one-flag change to the verify step (Section~\ref{sec:design}).
  \item \textbf{Two implementations}: in \texttt{vllm-torchtpu}~\cite{vllm_torchtpu,gaad_pr}, composing with MTP, DFlash and n-gram drafters (Section~\ref{sec:impl}), and in an independent serving runtime written in Rust, with MTP.\footnote{The Rust runtime is internal and not publicly released at the time of writing; we use it only to corroborate results obtained in \texttt{vllm-torchtpu}.}
  \item \textbf{An evaluation} of GaaD against the best configuration of each drafter alone across batch sizes, with ablations that separate the grammar drafts from the wider row, reproduced in both implementations, and of its cost to plain-text requests under mixed traffic (Section~\ref{sec:eval}).
  \item \textbf{An analysis of tokenization drift}: why committing grammar drafts can change greedy output, confirmed against an fp32 reference (Sections~\ref{sec:semantics} and~\ref{sec:equivalence}).
\end{enumerate}

\section{Background}

\textbf{Grammar-masked decoding.} A grammar engine such as XGrammar~\cite{dong2025xgrammar} keeps a matcher per request and, before each sampling step, computes a bitmask of the tokens the grammar allows next. The engine sets the logits of all other tokens to $-\infty$ and samples. XGrammar can also return the \emph{jump-forward string}: the longest string the grammar forces from the matcher's current state.

\textbf{Speculative decoding.} A drafter proposes up to $K$ tokens; the target model scores all of them in one forward pass over $K{+}1$ positions, accepts a prefix of them, and samples one more token itself; rejection sampling keeps the output distribution that of the target~\cite{leviathan2023fast,chen2023accelerating}. Under greedy decoding the accepted prefix is the longest one that matches the target's own choices, and the output is identical to decoding without drafts. Drafters range from host-side n-gram lookup~\cite{saxena2023promptlookup} to small models run on the accelerator~\cite{li2025eagle3,gloeckle2024better,dflash}. With structured outputs, vLLM masks the target's logits at each draft position with the bitmask the grammar gives after the preceding drafts, so the verify step enforces the grammar too~\cite{vllm_structured_output}.

\textbf{Static shapes on TPU.} \texttt{vllm-torchtpu} compiles the model ahead of time for a fixed set of token-count buckets (\emph{bucketization}) and pads each step to the nearest bucket. A verify step flattens every request's draft row, up to $K$ drafts plus the bonus position, into one token batch padded to a bucket, so rows of any length up to $K$ reuse the same precompiled shapes with no extra forward pass. By contrast, a separate variable-length jump-forward prefill either adds new compiled shapes or pays for an extra padded forward pass on the critical path.

\section{Design}
\label{sec:design}

\subsection{Grammar tokens as draft slots}

For each running request, the scheduler builds a draft row of at most $K$ tokens. As in speculative decoding generally, $K$ is the number of draft tokens a row holds; we write $K_M \le K$ for the model drafter's share, and without GaaD $K = K_M$. Each slot carries a flag: \texttt{MODEL} for a token proposed by a drafter, \texttt{GRAMMAR} for a token the grammar forces. The verify step treats \texttt{MODEL} slots as usual and accepts a \texttt{GRAMMAR} slot without comparing it with the target's choice. The first rejected slot still rejects every slot after it. Because \texttt{GRAMMAR} slots are accepted without verification, the verify step only needs their forward pass to populate the KV cache and skips their LM-head computation.

\subsection{The grammar's own draft}

From the current grammar state, GaaD builds the grammar's draft in two steps: (1)~query the grammar engine for the longest \emph{jump-forward string} uniquely forced before the next free choice, encode it with the canonical tokenization, and check that the grammar accepts every token in order (without retokenizing prior history; if nothing is forced or any token is rejected, the draft is empty); and (2)~truncate the verified tokens to at most $K$ slots. Although a forced string may admit multiple tokenizations, GaaD always drafts this canonical one (Section~\ref{sec:semantics}).

\subsection{Combining with a model drafter}

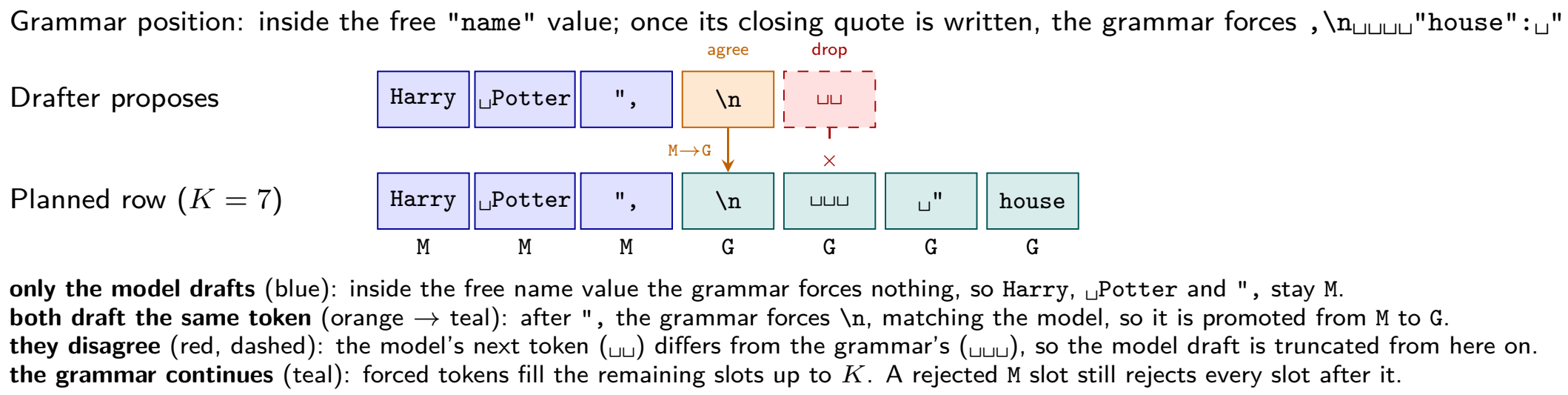
\begin{figure*}[t]
  \centering
  \resizebox{0.8\textwidth}{!}{\input{figures/fig3_planner.tex}}
  \caption{Illustrative example of combining the grammar's draft with a model drafter's (Qwen tokenization, $K_M{=}5$, $K{=}7$). Model tokens at free positions are kept as \texttt{M}, those that agree with a forced token as \texttt{G}; the model draft stops where the two disagree, and forced tokens fill the remaining slots.}
  \label{fig:planner}
\end{figure*}

A model drafter (MTP, DFlash or n-gram) proposes up to $K_M \le K$ tokens (its maximum proposal length) without seeing the grammar. The planner merges the two drafts position by position, starting from the request's committed grammar state; at each position it asks the grammar, from the state reached so far, whether it forces a token there (Figure~\ref{fig:planner}):
\begin{enumerate}
  \item \textbf{Both draft the same token.} At a forced position, if the model token equals the forced token, it is kept and flagged \texttt{GRAMMAR}.
  \item \textbf{Only the model drafts.} At a free position the grammar forces no token. Here a model token allowed by the grammar is kept and flagged \texttt{MODEL}, and the verify step checks it as usual.
  \item \textbf{Drafts disagree.} At the first model token that differs from the grammar's token, or that the grammar does not allow, the planner truncates the model draft early and drops all remaining model tokens (which were conditioned on that discarded token anyway). This early truncation avoids wasting verify-step compute on invalid or mismatched model slots and frees the remaining row capacity for \texttt{GRAMMAR} tokens whenever the grammar forces a continuation.
  \item \textbf{The grammar continues.} After the last kept model token, the planner keeps asking the grammar from the state reached there and appends each forced token, flagged \texttt{GRAMMAR}, until the grammar forces nothing or the row has $K$ slots.
\end{enumerate}
A row of $K{=}8$ with an MTP drafter proposing $K_M{=}3$ costs the drafter nothing extra, and the verify step computes logits only for the model slots and the bonus slot. During planning, the planner speculatively advances the grammar matcher over each planned token and rolls it back afterwards; the committed grammar state advances only over the tokens accepted in the verify step.

Because the model drafter proposes its tokens during step $t$ (on device for MTP and DFlash) before the host computes the grammar draft, both drafts start at the same position. The planner therefore cannot shift model tokens past a forced string---drafting \emph{after} a forced string would require an extra drafter pass on the critical path. Instead, whenever the drafter already predicts the forced tokens (common on fixed syntax), case~1 promotes them to \texttt{GRAMMAR} and keeps the drafter's subsequent tokens for free.

\subsection{Synchronous row planning}
\label{sec:sync}

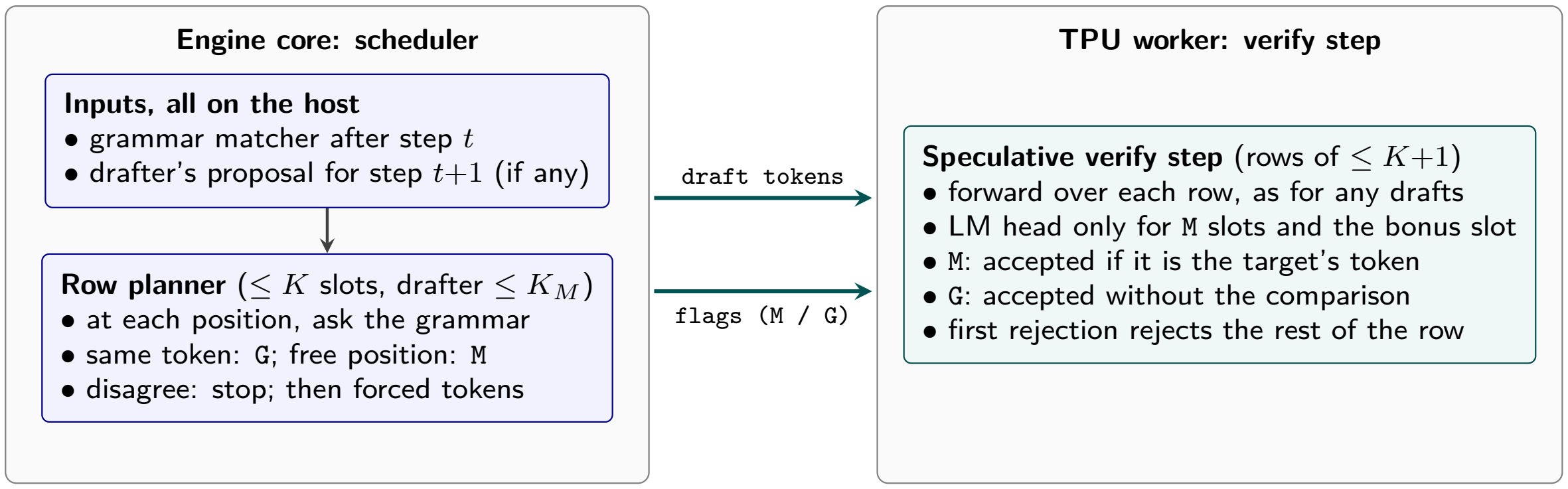
\begin{figure*}[t]
  \centering
  \resizebox{0.88\textwidth}{!}{\input{figures/fig2_architecture.tex}}
  \caption{GaaD in the engine. The scheduler plans each row from the grammar state and the drafter's proposal before the step is dispatched; the worker runs the verify step on token buckets compiled at start-up and accepts \texttt{G} slots.}
  \label{fig:arch}
\end{figure*}

To compute the jump-forward string for step $t{+}1$, the host needs the grammar state after step $t$'s accepted tokens---which are only available once the device finishes step $t$ and returns them to the host. With GaaD enabled, step $t{+}1$ therefore cannot overlap with step $t$ under asynchronous scheduling: the host waits for step $t$'s accepted tokens, advances the grammar matcher, plans the rows of step $t{+}1$, and only then dispatches it, leaving the device idle while the host works (Figure~\ref{fig:arch}).

Because whether step $t{+}1$ will hit a forced string is unknown until step $t$'s tokens arrive, both runtimes decide per batch: any batch containing a structured request runs synchronously (including plain-text requests sharing that batch), while plain-only batches stay asynchronous when using a device drafter (MTP or DFlash). With host-side n-gram drafting or GaaD alone, \texttt{vllm-torchtpu} disables asynchronous scheduling engine-wide. In the Rust runtime with a device drafter, every batch with a structured request already runs synchronously even without GaaD, so GaaD adds no extra synchronous steps there. Sections~\ref{sec:process} and~\ref{sec:mixed} measure this synchronization cost, including its impact on plain-text requests under mixed traffic.

\subsection{What grammar drafts change}
\label{sec:semantics}

A grammar constrains text, whereas the model generates tokens, and the same forced string can admit multiple valid tokenizations. Under a grammar mask, greedy decoding picks the highest-logit valid prefix token, which need not match the tokenizer's canonical encoding. Because GaaD commits canonical tokens for \texttt{GRAMMAR} slots (except when a row ends mid-string and the target samples the bonus token under the mask), every subsequent step is conditioned on the canonical tokenization instead of the model's own choice.

This choice can be fragile. For Qwen3.5-0.8B on our workload, an fp32 CPU reference (Hugging Face Transformers, Section~\ref{sec:equivalence}) scores a single space barely above the canonical three spaces after \texttt{\{\textbackslash n} (logits 16.88 vs.\ 16.63, a margin of 0.25; Qwen's tokenizer merges the fourth space with the quote into \texttt{\textvisiblespace"}). Greedy decoding picks the single space and continues that finer tokenization (four single-space tokens, \texttt{"house"} as \texttt{"h}\,$|$\,\texttt{ous}\,$|$\,\texttt{e}), emitting about 40\% more tokens overall; a second runtime whose arithmetic tips that 0.25 margin produces the canonical tokenization from the start (Section~\ref{sec:rust}).

Both tokenizations spell valid text, so neither is an error; in fact, the canonical tokenization matches the tokenizer's encoding of training text and is more compact. However, accepting canonical \texttt{GRAMMAR} slots without verification replaces the target's distribution over tokenizations (and the values conditioned on it), which is where GaaD departs from lossless speculative decoding. We call this effect \emph{tokenization drift}: GaaD and the drafter alone can produce different valid outputs, and because canonical tokenization uses fewer tokens, part of GaaD's speedup comes from it.

\section{Implementation in vLLM on TPU}
\label{sec:impl}

We implement GaaD in \texttt{vllm-torchtpu}~\cite{gaad_pr}, the TPU backend of vLLM, on top of vLLM 0.33. The implementation touches the scheduler, the engine's step loop, the speculative-decoding metadata, and the rejection sampler; it adds no kernel, and the graphs it uses are compiled at start-up rather than on the critical path.

\textbf{Scheduler.} The scheduler plans every running request's row before it allocates tokens, and the row replaces the request's draft tokens. Requests are skipped while they are still prefilling, while a reasoning parser reports that the grammar does not yet apply, when they report log-probabilities or apply presence, frequency or repetition penalties, which we leave to drafters alone, or when the grammar has terminated. The scheduler sends the flags of each scheduled row to the worker in the scheduler output. The forced tokens reach the model through the same path as any draft tokens.

\textbf{Synchronous steps.} With MTP or DFlash, asynchronous scheduling stays on. The engine commits the step in flight before scheduling the next one whenever a running request is eligible for GaaD, so the scheduler plans that step from committed grammar state and the drafter's proposals on the host; every other step stays asynchronous (Section~\ref{sec:sync}). With n-gram or GaaD alone there are no device drafts to feed asynchronous steps, and asynchronous scheduling is off.

\textbf{Drafter share.} The speculative-decoding configuration sets the total row width $K$, while GaaD's configuration sets the drafter's share $K_M$; EAGLE-3 and MTP drafters run only $K_M$ draft steps, and every request's drafts are cut to $K_M$ before planning.

\textbf{Verify step.} The worker computes target logits only for \texttt{MODEL} drafts and bonus slots, except in steps whose requests need every verify row, such as those with a minimum output length. It passes each draft's flag to both the greedy and random rejection samplers, which accept a \texttt{GRAMMAR} draft without their usual test. Fewer logit rows map to smaller token buckets for the LM head and logits gather, which the engine warms at start-up along with the samplers' flagged graph, so GaaD adds no compilation on the critical path.

\section{Evaluation}
\label{sec:eval}

\subsection{Setup}

\textbf{Implementations.} We evaluate GaaD in two serving engines. \texttt{vllm-torchtpu}~\cite{vllm_torchtpu} is vLLM's TPU backend, written in Python on TorchTPU (\texttt{torch.compile}) with JAX/Pallas kernels, with GaaD as described in Section~\ref{sec:impl}. The second is a serving runtime written in Rust that executes ahead-of-time exported StableHLO programs through the PJRT C API; it implements the same planning rule, row width, drafter share and LM-head skip, and runs a step synchronously whenever its batch holds a structured request. The two share the model weights, the tokenizer, XGrammar, and the workload, but differ in everything else, so a result that holds in both is a property of GaaD rather than of one engine.

\textbf{Platform.} One TPU v7x slice of four chips (eight TensorCores, topology 2x2x1) for both engines. Tensor parallelism is 1 for Qwen3.5-0.8B and Qwen3-4B, 4 for Qwen3.5-35B-A3B and 8 for Qwen3.5-397B-A17B-FP8.

\textbf{Models and drafters.} MTP with Qwen3.5-0.8B and Qwen3.5-35B-A3B (\texttt{vllm-torchtpu}), and with Qwen3.5-0.8B and Qwen3.5-397B-A17B-FP8 (Rust runtime); DFlash~\cite{dflash} (the z-lab checkpoint, block size 16) and n-gram with Qwen3-4B~\cite{yang2025qwen3} (\texttt{vllm-torchtpu}). Each drafter alone uses its usual $K = K_M$: 3 for MTP and 7 for DFlash and n-gram. With GaaD, in both engines, $K{=}8$ and the drafter keeps its $K_M$, as in the Rust runtime's default.

\textbf{Workload.} The JSON character-profile benchmark~\cite{yin2024compressedfsm}: 50 prompts, cycled to 256 requests for Table~\ref{tab:drafters}, 64 for the output comparison and 128 in the Rust runtime, each answered under a regular expression for a JSON object with eleven fixed keys (one of them, \texttt{wand}, holds a nested object) and fixed indentation, whose ten value fields are four enumerations, one number and five free-text strings; greedy decoding, at most 256 output tokens, prefix caching off.

\textbf{Arms.} We compare no-GaaD against GaaD on a best-against-best basis. The baseline is each drafter alone at its standard $K = K_M$, using the faster of synchronous and asynchronous scheduling at each batch size (n-gram drafts on the host and runs synchronously only; in the Rust runtime, default scheduling is already synchronous on structured batches). GaaD keeps the same $K_M$, widens the row to $K{=}8$, and runs synchronously on structured batches (Section~\ref{sec:sync}). Two ablations isolate the two changes: the drafter alone with $K{=}8$ (MTP, asynchronous), and GaaD constrained to $K = K_M$.

\textbf{Metrics.} We report two metrics. \emph{Average acceptance length} $\tau$~\cite{li2024eagle} is total output tokens divided by total target forward passes---equivalently, the average number of accepted draft tokens per verify step plus the bonus token. \emph{Speedup} is the baseline's end-to-end wall time divided by GaaD's on the same request set at the same maximum concurrency $B$ (measured after warm-up; median of multiple passes in \texttt{vllm-torchtpu}).

\subsection{GaaD with model drafters}
\label{sec:drafters}

\begin{table*}[t]
  \centering
  \caption{GaaD with model drafters in \texttt{vllm-torchtpu} (256 requests): average acceptance length $\tau$ for the drafter alone ($K = K_M$) versus with GaaD ($K{=}8$), and end-to-end speedup. \emph{Best vs.\ best}: GaaD ($K{=}8$) over the best drafter-alone baseline ($K = K_M$, faster of sync/async). \emph{vs.\ $K{=}8$}: GaaD ($K{=}8$) over MTP alone widened to $K{=}8$ (async). \emph{GaaD ($K{=}K_M$)}: GaaD restricted to $K = K_M$ (grammar drafts without a wider row) over the best drafter-alone baseline.}
  \label{tab:drafters}
  \small
  \input{tables/drafters.tex}
\end{table*}

\begin{figure}[t]
  \centering
  \input{figures/fig4_batch_scaling.tex}
  \caption{Speedup of GaaD ($K{=}8$) over the best configuration of the drafter alone at each point, in \texttt{vllm-torchtpu}.}
  \label{fig:sweep}
\end{figure}
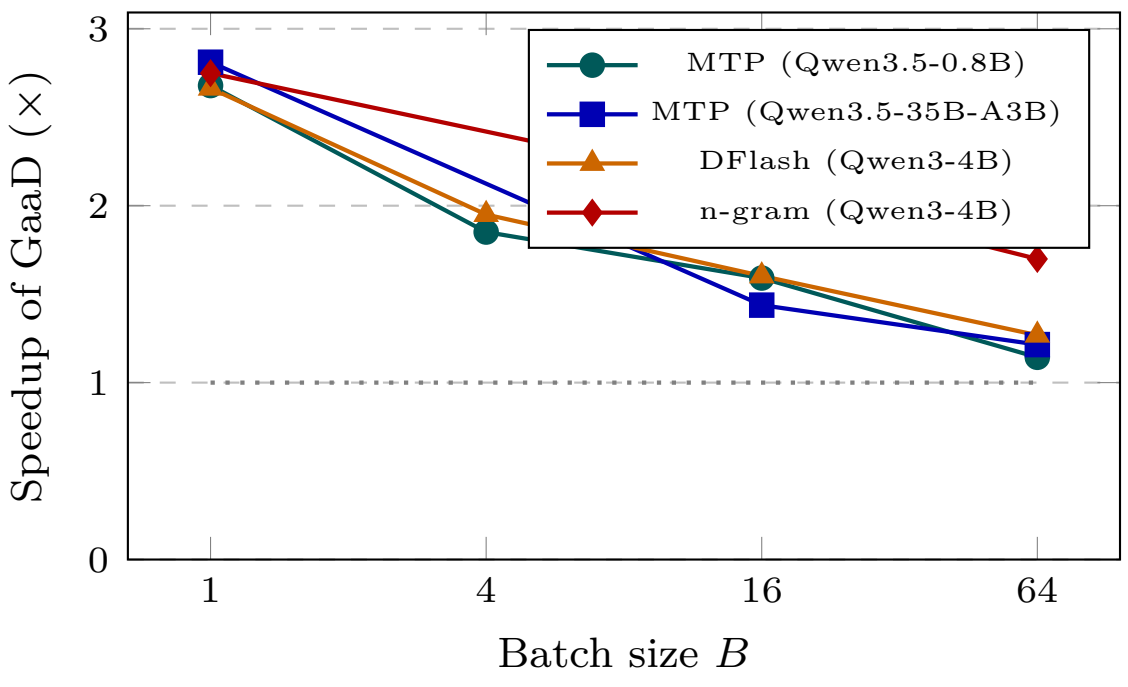

Table~\ref{tab:drafters} is the main result. With only the drafters, acceptance length reaches $\tau$ = 1.30--2.06. With GaaD the drafter keeps its $K_M$ slots and the grammar fills the rest of a row of 8: $\tau$ rises to 5.26--5.43 with MTP, 4.11--4.13 with DFlash and 3.61--3.64 with n-gram, at every batch size. Against the best configuration of each drafter alone, decoding is 2.68$\times$ (MTP, Qwen3.5-0.8B), 2.81$\times$ (MTP, Qwen3.5-35B-A3B), 2.66$\times$ (DFlash) and 2.75$\times$ (n-gram) faster at $B{=}1$, 1.44--2.09$\times$ faster at $B{=}16$ and 1.14--1.70$\times$ faster at $B{=}64$ (Figure~\ref{fig:sweep}). Wall time is computed as the number of output tokens divided by $\tau$, and multiplied by the wall time each request step takes (a request step is one verify step of one request). So the speedup is the product of three measured factors (Table~\ref{tab:decomp}): the reference's output tokens over GaaD's, GaaD's $\tau$ over the reference's, and the reference's wall time per request step over GaaD's. The first comes from tokenization (Section~\ref{sec:semantics}) and is 1.31--1.38 at every point: GaaD's outputs are as long as the drafter alone's in characters (within 1\%) but use fewer tokens, because the drafter alone's greedy output uses a finer tokenization of the same kind of text. The second is 2.62--2.81, also at every batch size. The third is below one everywhere, and the whole decline with $B$ is in it (Section~\ref{sec:process}). In the Rust runtime, where both arms tokenize canonically and emit about as many tokens, MTP with GaaD is still 2.02--2.23$\times$ faster than MTP alone at concurrency 1 (Section~\ref{sec:rust}). Section~\ref{sec:rust} shows the same trend in a second engine, and Section~\ref{sec:process} examines why the gain shrinks with $B$.

The ablations separate GaaD's two changes. A wider row alone does not help: under the same asynchronous scheduling, MTP alone with $K{=}8$ is 0.81--0.95$\times$ as fast as with $K{=}3$ on Qwen3.5-0.8B and 0.67--0.89$\times$ on Qwen3.5-35B-A3B (with $K{=}7$, synchronous, 0.84--0.98$\times$ on Qwen3.5-0.8B), so against MTP alone at the same $K{=}8$, GaaD is 1.49--3.58$\times$ faster. Grammar drafts without a wider row help less: GaaD with $K = K_M$ is 1.03--2.52$\times$ faster than the best drafter alone, and GaaD with $K{=}8$ is faster than with $K = K_M$ at every batch size.

\begin{table*}[t]
  \centering
  \caption{Factor decomposition of the best-vs.-best speedup of Table~\ref{tab:drafters} at the endpoints $B{=}1$ and $B{=}64$ (intermediate batch sizes follow the same monotonic trend): reference-to-GaaD output tokens (\emph{Tokens}), GaaD-to-reference acceptance length ($\tau$), and reference-to-GaaD wall time per request step (\emph{Per step}), whose product equals \emph{Speedup}. \emph{Reference}: scheduling of the faster drafter-alone baseline.}
  \label{tab:decomp}
  \small
  \input{tables/decomposition.tex}
\end{table*}

\subsection{A second implementation in Rust}
\label{sec:rust}

\begin{table}[t]
  \centering
  \caption{Rust runtime: speedup of MTP + GaaD over MTP alone (MTP alone's wall time over MTP + GaaD's, for the same 128 requests) at each concurrency.}
  \label{tab:rust}
  \small
  \input{tables/rust.tex}
\end{table}

The Rust runtime reproduces the result in a second implementation (Table~\ref{tab:rust}): MTP with GaaD is 2.23$\times$ faster than MTP alone on Qwen3.5-0.8B and 2.02$\times$ on Qwen3.5-397B-A17B-FP8 at concurrency 1. The gain also shrinks as concurrency grows, but less steeply than in \texttt{vllm-torchtpu}: on the 0.8B model it is 1.37$\times$ at 64 concurrent requests and still 1.22$\times$ at 128, against 1.14$\times$ for MTP on the same model at $B{=}64$ in \texttt{vllm-torchtpu}; on the 397B model it is 1.69$\times$ at 32. Unlike \texttt{vllm-torchtpu}, MTP alone and MTP with GaaD in the Rust runtime emit within 0.3\% as many output tokens at every concurrency, and at concurrency 1 produce identical text for all 128 requests on 0.8B and 82 of 128 on 397B, so the speedup here comes from acceptance length rather than shorter tokenization.

\subsection{Output equivalence and tokenization drift}
\label{sec:equivalence}

Across all eight drafter, model, and batch-size configurations in \texttt{vllm-torchtpu} ($B \in \{1, 16\}$, 64 requests), GaaD matches 0/64 outputs token-for-token against the drafter alone and emits 24--28\% fewer tokens. This is not batching noise: at $B{=}1$ the drafter alone produces identical outputs under both schedulers (64/64), and differences between schedulers at $B{=}16$ (5--48 of 64 identical) begin inside model-chosen values, where bfloat16 arithmetic flips near-ties.

The difference is the tokenization drift of Section~\ref{sec:semantics}, which we confirm with an fp32 reference: Hugging Face Transformers on CPU, decoding Qwen3.5-0.8B greedily under the same grammar mask. For the first three prompts it produces, token for token, the finer-grained outputs \texttt{vllm-torchtpu} produces without GaaD; the margin between the two best allowed tokens is 0.25, 0.72 and 0.69 at the first indentation and 7.5--9.1 at the next two choices. \texttt{vllm-torchtpu}'s baseline is therefore the model's own greedy output, and GaaD's shorter outputs are the canonical tokenization of the same syntax.

Overall the two outputs compare as follows. Every output of every arm matches the regular expression. GaaD and the drafter alone choose the same value for 85--96\% of the enumeration fields, but for only 47--69\% of the numbers and 50--74\% of the free-text fields: once the tokenization differs, later choices differ too. Under the model itself, neither side is systematically more likely. Rescored by the fp32 reference with both texts encoded canonically, GaaD's text is the more likely one for 31 of 64 requests with Qwen3.5-0.8B (MTP) and for 31 of 64 with Qwen3-4B (n-gram), and the median per-request difference is zero for both; the mean log-likelihood is $-178.58$ against $-178.51$ for Qwen3.5-0.8B and $-137.51$ against $-135.03$ for Qwen3-4B, where a few requests favour the drafter alone. We do not measure task accuracy.

\subsection{What limits large batches}
\label{sec:process}

Why does the gain shrink with batch size? Not because the grammar drafts stop being accepted---$\tau$ stays at 5.26--5.43 with MTP, 4.11--4.13 with DFlash and 3.61--3.64 with n-gram at every $B$ in Table~\ref{tab:drafters}. Table~\ref{tab:decomp} locates the decline: the tokenization and $\tau$ factors do not change with $B$, while the third factor, wall time per request step, falls from 0.73--0.80 at $B{=}1$ to 0.31--0.46 at $B{=}64$. Two main effects drive this decline:

\begin{enumerate}
  \item \textbf{Wider verify rows.} Once the batch is large enough that the verify step is no longer memory-bandwidth bound, step time grows with the positions processed: MTP alone with $K{=}8$ is only 0.81$\times$ as fast as with $K{=}3$ at $B{=}64$ on Qwen3.5-0.8B, and 0.67$\times$ on Qwen3.5-35B-A3B. GaaD's grammar drafts still more than recover that cost (GaaD with $K{=}8$ beats $K = K_M$ at every $B$), but the net margin narrows with $B$. This holds in both engines (including Rust, Section~\ref{sec:rust}), though it hardly affects DFlash and n-gram, whose rows grow only from 8 to 9 positions.
  \item \textbf{Synchronous host overhead.} Because every batch on this workload holds a structured request, \texttt{vllm-torchtpu} runs every GaaD step synchronously (Section~\ref{sec:sync}), exposing per-step Python scheduler and cross-process host work that grows with $B$. The Rust runtime is synchronous too but has no Python scheduler or process boundary, so it retains more speedup at $B{=}64$ on Qwen3.5-0.8B with MTP: 1.37$\times$ (Table~\ref{tab:rust}) versus 1.14$\times$ in \texttt{vllm-torchtpu}.
\end{enumerate}

At $B{=}64$, part of the remaining per-request-step gap may also come from lower average batch fill near the tail of a run if GaaD shortens average requests more than the longest free-text-heavy requests; separating per-step time from batch fill requires counting engine steps (Section~\ref{sec:future}).

\subsection{Mixed traffic}
\label{sec:mixed}

Because serving traffic mixes structured and unconstrained requests, we test whether GaaD hurts plain-text requests sharing an engine (Qwen3.5-0.8B with MTP, varying the structured share in \texttt{vllm-torchtpu} and keeping $N$ structured requests in flight among $c$ plain requests in the Rust runtime). Three findings hold across both engines. First, on \textbf{plain-only traffic} (0\% structured), per-batch synchronization keeps plain requests asynchronous; in the Rust runtime plain TPOT is unchanged (1.00--1.01$\times$), and in \texttt{vllm-torchtpu} plain requests maintain the exact same acceptance length ($\tau = 3.49$--$3.51$). Second, on \textbf{mixed traffic} (25--50\% structured), plain requests in shared batches pay a modest TPOT penalty (+7--42\% in \texttt{vllm-torchtpu}, +7--47\% in Rust, on top of the slowdown structured requests already cause without GaaD), while structured requests finish 2.2--3.2$\times$ faster in \texttt{vllm-torchtpu} (1.59--1.88$\times$ lower TPOT in Rust). Third, \textbf{aggregate performance improves or holds}: total wall time is 1.03--1.23$\times$ faster in \texttt{vllm-torchtpu} under per-batch synchronization, and total output throughput stays within 0.94--1.03$\times$ in Rust. In short, GaaD shifts step time from plain requests to structured ones within a shared batch rather than adding net overhead.

\subsection{Host cost}
\label{sec:host}

\begin{table}[t]
  \centering
  \caption{Host time GaaD spends planning rows per engine step (MTP + GaaD on Qwen3.5-0.8B, $K{=}8$, $K_M{=}3$, 256 requests, single-process engine), and its share of the engine step's wall time, which includes prefill and the single-process engine's synchronous overhead.}
  \label{tab:host}
  \small
  \input{tables/host.tex}
\end{table}

Planning itself is cheap (Table~\ref{tab:host}): 0.136\,ms per engine step at $B{=}1$, 0.91\,ms at $B{=}16$ and 2.74\,ms at $B{=}64$, or 1.0\%, 1.3\% and 1.2\% of the step.

\section{Related Work}

\textbf{Constrained decoding.} Outlines~\cite{willard2023efficient} and SynCode~\cite{ugare2024syncode} compile grammars to automata and mask logits per step; XGrammar~\cite{dong2025xgrammar} makes the mask nearly free and provides the jump-forward strings GaaD uses. DOMINO~\cite{beurerkellner2024domino} aligns grammar constraints with subword tokens and uses speculative lookahead to make constrained decoding cheap; it argues that constraining text without regard to tokenization distorts the model's output, which our tokenization-drift measurements quantify for grammar drafts.

\textbf{Jump-forward and fast-forward tokens.} SGLang's compressed FSM~\cite{yin2024compressedfsm,zheng2024sglang} introduced jump-forward decoding with a separate extend step; coalescence in Outlines~\cite{kurt_coalescence} merges deterministic paths of the token automaton, and notes that the tokenization chosen for a forced string changes the model's later distribution---the effect we measure as tokenization drift; llguidance~\cite{llguidance} returns fast-forward tokens to the engine. GaaD differs in where the forced tokens go: into the speculative draft row, merged with a drafter's proposal and processed by the existing verify step.

\textbf{Speculative decoding.} Draft models~\cite{leviathan2023fast,chen2023accelerating}, Medusa~\cite{cai2024medusa}, EAGLE~\cite{li2024eagle,li2025eagle3}, MTP~\cite{gloeckle2024better}, DFlash~\cite{dflash}, SpecInfer~\cite{miao2024specinfer}, lookahead~\cite{fu2024break} and prompt-lookup~\cite{saxena2023promptlookup} decoding propose tokens from a model or from the context. GaaD adds a symbolic source whose tokens the verify step accepts without computing their logits.

\textbf{Serving systems.} vLLM~\cite{kwon2023efficient,vllm_structured_output} and TensorRT-LLM~\cite{nvidia_trtllm} support structured outputs together with speculative decoding; vLLM masks the verify step's logits at every draft position. GaaD reuses that verify step as the jump-forward mechanism.

\section{Limitations}
\label{sec:limits}

\begin{itemize}
  \item \textbf{Mixed traffic.} Batches containing a structured request run synchronously (Section~\ref{sec:sync}), raising plain-text TPOT in shared batches by 7--42\% in \texttt{vllm-torchtpu} and 7--47\% in the Rust runtime while accelerating structured requests by 1.6--3.2$\times$ (Section~\ref{sec:mixed}). Deployments that prioritize plain-text tail latency over structured latency must account for this trade-off.
  \item \textbf{Not lossless.} Because GaaD accepts canonical \texttt{GRAMMAR} drafts with probability 1, it changes token boundaries at forced positions and is not token-for-token equivalent to grammar-masked decoding (Section~\ref{sec:semantics}). Under greedy decoding, every output remains grammar-valid and an fp32 reference rates both outputs equally likely (Section~\ref{sec:equivalence}), but we do not evaluate random sampling or downstream task accuracy.
  \item \textbf{Evaluation scope.} Our experiments cover one workload (the JSON character-profile benchmark), greedy decoding, and models from 0.8B to 397B parameters on a single TPU v7x slice. This schema fixes a large fraction of each output (eleven keys, indentation, and delimiters); schemas dominated by long free-text fields leave fewer tokens for the grammar to draft.
\end{itemize}

\section{Discussion and Future Work}
\label{sec:future}

\textbf{Asynchronous grammar drafting.} Host-side synchronization is GaaD's largest remaining overhead at high concurrency and under mixed traffic (Sections~\ref{sec:sync} and~\ref{sec:mixed}). Because a grammar draft depends only on the previous step's accepted tokens, moving lightweight deterministic lookahead closer to the verify step---or isolating structured batches from unconstrained requests in the scheduler---would remove the synchronization penalty for both structured and plain traffic.

\textbf{Batch-adaptive row width $K$.} Wider rows pay off most in the memory-bandwidth-bound regime (small $B$) and cost more as the verify step becomes compute-bound (Section~\ref{sec:process}). Dynamically scheduling $K$ by batch size---wide at low concurrency and narrower at high concurrency---can preserve large-batch throughput while keeping low-latency gains.

\textbf{Distribution-preserving sampling and richer deterministic sources.} Under non-greedy sampling, accepting a single canonical tokenization alters the token-level distribution; designing a rejection rule that aggregates probability mass across valid tokenizations of a forced string remains open. Finally, the same one-flag \texttt{GRAMMAR} slot generalizes beyond regex and JSON grammars to any deterministic span---such as fixed tool-call templates, API call skeletons, or constrained prompt copies---without changing the verify kernel.

\section{Conclusion}

On structured outputs, a speculative drafter wastes much of its row on tokens the grammar already knows. Grammar-as-a-Drafter writes those tokens into the row itself, merged with the drafter's proposal, in a row wider than the drafter fills, and accepted without computing their logits. On the JSON character-profile benchmark in \texttt{vllm-torchtpu}, it raises the average acceptance length to 3.6--5.4 for MTP, DFlash, and n-gram drafters and decodes up to 2.8$\times$ faster than the best configuration of each drafter alone; a second, independent Rust runtime reproduces the speedup with MTP up to 397B parameters (2.0--2.2$\times$ at concurrency 1, and 1.22--1.69$\times$ at high concurrency). Under mixed traffic it speeds up structured requests at a modest TPOT cost to plain-text requests sharing a batch, while improving or preserving aggregate throughput. Because grammars fix text rather than tokens, accepted grammar drafts can change the output where greedy decoding chooses a different tokenization of forced text---a property of grammar-masked decoding that an fp32 reference confirms, and one engines should account for. For structured outputs, the grammar is the cheapest drafter an engine has.

\bibliographystyle{plain}
\bibliography{references}

\end{document}

%% file: figures/fig1_conflicts.tex
\begin{tikzpicture}[
  font=\footnotesize\sffamily,
  box/.style={draw, rounded corners=2pt, minimum height=6.5mm, inner xsep=5pt, align=center, text width=3.7cm},
  prior/.style={box, draw=red!65!black, fill=red!6},
  gaad/.style={box, draw=teal!65!black, fill=teal!8},
  arr/.style={->, >=stealth, thick, draw=black!65}
]
  \node[anchor=west, font=\footnotesize\bfseries\sffamily, text=red!65!black] at (0, 1.35)
    {Jump-forward as a micro-prefill};
  \node[prior] (p1) at (2.15, 0.75) {Grammar forces\\$M$ tokens};
  \node[prior] (p2) at (6.75, 0.75) {Extra forward of\\ragged shape $[1, M]$};
  \node[prior] (p3) at (11.35, 0.75) {New compiled shape or\\padding to a bucket};
  \node[prior] (p4) at (15.95, 0.75) {Model drafts for the\\same positions discarded};
  \draw[arr] (p1) -- (p2);
  \draw[arr] (p2) -- (p3);
  \draw[arr] (p3) -- (p4);

  \node[anchor=west, font=\footnotesize\bfseries\sffamily, text=teal!65!black] at (0, -0.15)
    {Jump-forward as draft tokens (GaaD)};
  \node[gaad] (g1) at (2.15, -0.75) {Grammar forces\\$M$ tokens};
  \node[gaad, font=\footnotesize\ttfamily] (g2) at (6.75, -0.75) {[ G G G M G G ]};
  \node[gaad] (g3) at (11.35, -0.75) {Verify step on token buckets\\compiled at start-up};
  \node[gaad] (g4) at (15.95, -0.75) {Model drafts kept where\\they agree with the grammar};
  \draw[arr, draw=teal!65!black] (g1) -- (g2);
  \draw[arr, draw=teal!65!black] (g2) -- (g3);
  \draw[arr, draw=teal!65!black] (g3) -- (g4);
\end{tikzpicture}

%% file: figures/fig3_planner.tex
\begin{tikzpicture}[
  font=\footnotesize\sffamily,
  slot/.style={draw=black!60, minimum width=9mm, minimum height=5.5mm, inner sep=1pt, font=\scriptsize\ttfamily, anchor=center},
  gslot/.style={slot, fill=teal!15, draw=teal!65!black},
  mslot/.style={slot, fill=blue!12, draw=blue!60!black},
  aslot/.style={slot, fill=orange!18, draw=orange!75!black},
  dslot/.style={slot, fill=red!12, draw=red!65!black, dashed, text=red!60!black},
  lbl/.style={anchor=west, font=\footnotesize\sffamily}
]
  \node[lbl] at (0, 2.7) {Grammar position: inside the free \texttt{"name"} value; once its closing quote is written, the grammar forces \texttt{,\textbackslash n\textvisiblespace\textvisiblespace\textvisiblespace\textvisiblespace"house":\textvisiblespace"}};

  \node[lbl] at (0, 1.95) {Drafter proposes};
  \foreach \i/\t/\st in {0/Harry/mslot, 1/{\textvisiblespace Potter}/mslot, 2/{",}/mslot, 3/{\textbackslash n}/aslot, 4/{\textvisiblespace\textvisiblespace}/dslot} {
    \node[\st] (d\i) at (4.2 + \i*1.0, 1.95) {\t};
  }
  \node[font=\tiny\sffamily, text=orange!75!black, anchor=south] at (d3.north) {agree};
  \node[font=\tiny\sffamily, text=red!65!black, anchor=south] at (d4.north) {drop};

  \node[lbl] at (0, 0.95) {Planned row ($K=7$)};
  \foreach \i/\t/\st in {0/Harry/mslot, 1/{\textvisiblespace Potter}/mslot, 2/{",}/mslot, 3/{\textbackslash n}/gslot, 4/{\textvisiblespace\textvisiblespace\textvisiblespace}/gslot, 5/{\textvisiblespace"}/gslot, 6/{house}/gslot} {
    \node[\st] (p\i) at (4.2 + \i*1.0, 0.95) {\t};
  }
  \draw[->, >=stealth, orange!75!black, semithick] (d3.south) -- (p3.north) node[midway, left=1pt, font=\tiny\sffamily] {\texttt{M}$\to$\texttt{G}};
  \draw[red!65!black, dashed, semithick] (d4.south) -- ++(0,-0.16) node[below=-1pt, font=\tiny\sffamily] {$\times$};

  \foreach \i/\t in {0/M, 1/M, 2/M, 3/G, 4/G, 5/G, 6/G} {
    \node[font=\scriptsize\ttfamily] at (4.2 + \i*1.0, 0.5) {\t};
  }

  \node[anchor=west, font=\scriptsize\sffamily, align=left] at (0, -0.35) {
    \textbf{only the model drafts} (blue): inside the free name value the grammar forces nothing, so \texttt{Harry}, \texttt{\textvisiblespace Potter} and \texttt{",} stay \texttt{M}.\\
    \textbf{both draft the same token} (orange $\to$ teal): after \texttt{",} the grammar forces \texttt{\textbackslash n}, matching the model, so it is promoted from \texttt{M} to \texttt{G}.\\
    \textbf{they disagree} (red, dashed): the model's next token (\texttt{\textvisiblespace\textvisiblespace}) differs from the grammar's (\texttt{\textvisiblespace\textvisiblespace\textvisiblespace}), so the model draft is truncated from here on.\\
    \textbf{the grammar continues} (teal): forced tokens fill the remaining slots up to $K$. A rejected \texttt{M} slot still rejects every slot after it.
  };
\end{tikzpicture}

%% file: figures/fig2_architecture.tex
\begin{tikzpicture}[
  font=\footnotesize\sffamily,
  panel/.style={draw=black!50, rounded corners=3pt, fill=black!2, inner sep=6pt},
  subbox/.style={draw=blue!55!black, rounded corners=2pt, fill=blue!5, align=left, inner sep=5pt, minimum width=5.4cm},
  devbox/.style={draw=teal!60!black, rounded corners=2pt, fill=teal!6, align=left, inner sep=5pt, minimum width=5.8cm},
  arr/.style={->, >=stealth, thick, draw=black!75}
]
  \node[panel, minimum width=6.2cm, minimum height=4.6cm] (host) at (0, 0) {};
  \node[anchor=north, font=\footnotesize\bfseries\sffamily] at (host.north) [yshift=-3pt] {Engine core: scheduler};

  \node[subbox] (state) at (0, 1.0) {
    \textbf{Inputs, all on the host}\\
    $\bullet$ grammar matcher after step $t$\\
    $\bullet$ drafter's proposal for step $t{+}1$ (if any)
  };
  \node[subbox] (planner) at (0, -0.9) {
    \textbf{Row planner} ($\le K$ slots, drafter $\le K_M$)\\
    $\bullet$ at each position, ask the grammar\\
    $\bullet$ same token: \texttt{G}; free position: \texttt{M}\\
    $\bullet$ disagree: stop; then forced tokens
  };
  \draw[arr] (state) -- (planner);

  \node[panel, minimum width=6.6cm, minimum height=4.6cm] (dev) at (8.6, 0) {};
  \node[anchor=north, font=\footnotesize\bfseries\sffamily] at (dev.north) [yshift=-3pt] {TPU worker: verify step};

  \node[devbox] (graph) at (8.6, 0) {
    \textbf{Speculative verify step} (rows of $\le K{+}1$)\\
    $\bullet$ forward over each row, as for any drafts\\
    $\bullet$ LM head only for \texttt{M} slots and the bonus slot\\
    $\bullet$ \texttt{M}: accepted if it is the target's token\\
    $\bullet$ \texttt{G}: accepted without the comparison\\
    $\bullet$ first rejection rejects the rest of the row
  };

  \draw[arr, draw=teal!65!black, line width=1.1pt] (3.15, 0.45) -- node[above, font=\scriptsize\ttfamily] {draft tokens} (5.25, 0.45);
  \draw[arr, draw=teal!65!black, line width=1.1pt] (3.15, -0.45) -- node[below, font=\scriptsize\ttfamily] {flags (M / G)} (5.25, -0.45);
\end{tikzpicture}

%% file: tables/drafters.tex
\begin{tabular}{@{}llrrrrrr@{}}
\toprule
Drafter & Model & $B$ & \multicolumn{2}{c}{$\tau$} & \multicolumn{3}{c}{Speedup} \\
\cmidrule(lr){4-5}\cmidrule(lr){6-8}
 & & & alone & + GaaD & best vs.\ best & vs.\ $K{=}8$ & GaaD ($K{=}K_M$) \\
\midrule
MTP & Qwen3.5-0.8B & 1 & 2.06 & 5.43 & 2.68$\times$ & 3.39$\times$ & 1.85$\times$ \\
MTP & Qwen3.5-0.8B & 4 & 2.06 & 5.43 & 1.85$\times$ & 2.31$\times$ & -- \\
MTP & Qwen3.5-0.8B & 16 & 2.06 & 5.40 & 1.59$\times$ & 1.96$\times$ & 1.26$\times$ \\
MTP & Qwen3.5-0.8B & 64 & 2.05 & 5.41 & 1.14$\times$ & 1.49$\times$ & 1.03$\times$ \\
\midrule
MTP & Qwen3.5-35B-A3B & 1 & 1.99 & 5.31 & 2.81$\times$ & 3.58$\times$ & -- \\
MTP & Qwen3.5-35B-A3B & 16 & 1.99 & 5.29 & 1.44$\times$ & 2.00$\times$ & -- \\
MTP & Qwen3.5-35B-A3B & 64 & 1.99 & 5.26 & 1.21$\times$ & 1.81$\times$ & -- \\
\midrule
DFlash & Qwen3-4B & 1 & 1.57 & 4.13 & 2.66$\times$ & -- & 2.48$\times$ \\
DFlash & Qwen3-4B & 4 & 1.57 & 4.12 & 1.95$\times$ & -- & -- \\
DFlash & Qwen3-4B & 16 & 1.57 & 4.11 & 1.60$\times$ & -- & 1.46$\times$ \\
DFlash & Qwen3-4B & 64 & 1.57 & 4.12 & 1.27$\times$ & -- & 1.26$\times$ \\
\midrule
n-gram & Qwen3-4B & 1 & 1.30 & 3.64 & 2.75$\times$ & -- & 2.52$\times$ \\
n-gram & Qwen3-4B & 16 & 1.30 & 3.61 & 2.09$\times$ & -- & 1.81$\times$ \\
n-gram & Qwen3-4B & 64 & 1.30 & 3.62 & 1.70$\times$ & -- & 1.43$\times$ \\
\bottomrule
\end{tabular}

%% file: tables/decomposition.tex
\begin{tabular}{@{}llrlrrrr@{}}
\toprule
Drafter & Model & $B$ & Reference & Tokens & $\tau$ & Per step & Speedup \\
\midrule
MTP & Qwen3.5-0.8B & 1 & sync & 1.38 & 2.64 & 0.73 & 2.68$\times$ \\
MTP & Qwen3.5-0.8B & 64 & sync & 1.38 & 2.64 & 0.31 & 1.14$\times$ \\
\midrule
MTP & Qwen3.5-35B-A3B & 1 & sync & 1.32 & 2.66 & 0.80 & 2.81$\times$ \\
MTP & Qwen3.5-35B-A3B & 64 & async & 1.32 & 2.65 & 0.35 & 1.21$\times$ \\
\midrule
DFlash & Qwen3-4B & 1 & async & 1.31 & 2.62 & 0.77 & 2.66$\times$ \\
DFlash & Qwen3-4B & 64 & sync & 1.31 & 2.62 & 0.37 & 1.27$\times$ \\
\midrule
n-gram & Qwen3-4B & 1 & sync & 1.32 & 2.81 & 0.74 & 2.75$\times$ \\
n-gram & Qwen3-4B & 64 & sync & 1.31 & 2.79 & 0.46 & 1.70$\times$ \\
\bottomrule
\end{tabular}

%% file: figures/fig4_batch_scaling.tex
\begin{tikzpicture}
\begin{axis}[
  width=0.95\columnwidth, height=5.2cm,
  xlabel={Batch size $B$}, ylabel={Speedup of GaaD ($\times$)},
  xmode=log, log basis x=2, xtick={1,4,16,64}, xticklabels={1,4,16,64},
  ymajorgrids=true, grid style=dashed, ymin=0,
  legend pos=north east, legend style={font=\tiny},
  label style={font=\footnotesize}, tick label style={font=\scriptsize},
]
\addplot[color=teal!70!black, mark=*, thick] coordinates {(1, 2.679) (4, 1.851) (16, 1.591) (64, 1.143)};
\addlegendentry{MTP (Qwen3.5-0.8B)}
\addplot[color=blue!70!black, mark=square*, thick] coordinates {(1, 2.811) (16, 1.438) (64, 1.215)};
\addlegendentry{MTP (Qwen3.5-35B-A3B)}
\addplot[color=orange!80!black, mark=triangle*, thick] coordinates {(1, 2.664) (4, 1.950) (16, 1.603) (64, 1.268)};
\addlegendentry{DFlash (Qwen3-4B)}
\addplot[color=red!70!black, mark=diamond*, thick] coordinates {(1, 2.749) (16, 2.089) (64, 1.699)};
\addlegendentry{n-gram (Qwen3-4B)}
\addplot[color=black!50, dotted, thick, domain=1:64] {1.0};
\end{axis}
\end{tikzpicture}

%% file: tables/rust.tex
\begin{tabular}{@{}lrr@{}}
\toprule
Model & Conc. & Speedup \\
\midrule
Qwen3.5-0.8B & 1 & 2.23$\times$ \\
Qwen3.5-0.8B & 16 & 1.68$\times$ \\
Qwen3.5-0.8B & 64 & 1.37$\times$ \\
Qwen3.5-0.8B & 128 & 1.22$\times$ \\
\midrule
Qwen3.5-397B-A17B-FP8 & 1 & 2.02$\times$ \\
Qwen3.5-397B-A17B-FP8 & 8 & 1.86$\times$ \\
Qwen3.5-397B-A17B-FP8 & 32 & 1.69$\times$ \\
\bottomrule
\end{tabular}

%% file: tables/host.tex
\begin{tabular}{@{}rrr@{}}
\toprule
$B$ & Planning, ms/step & Share of step \\
\midrule
1 & 0.136 & 1.0\% \\
16 & 0.910 & 1.3\% \\
64 & 2.743 & 1.2\% \\
\bottomrule
\end{tabular}